\documentclass[10pt,twocolumn,letterpaper]{article}

\usepackage{cvpr}
\usepackage{times}
\usepackage{epsfig}
\usepackage{graphicx}
\usepackage{amsmath}
\usepackage{amssymb}
\usepackage{multirow}
\usepackage{array}
\usepackage{algorithm}
\usepackage{algorithmic}

\usepackage[breaklinks=true,bookmarks=false]{hyperref}

\cvprfinalcopy 

\def\cvprPaperID{1998} 

\ifcvprfinal\fi
\begin{document}

\title{AD-Cluster: Augmented Discriminative Clustering for Domain Adaptive \\Person Re-identification}

\author{Yunpeng Zhai$^{1,2}$, Shijian Lu$^{3}$, Qixiang Ye$^{4,6}$, Xuebo Shan$^{1,2}$, Jie Chen$^{1,6}$, \\Rongrong Ji$^{5,6}$, Yonghong Tian$^{1,2,6}$\thanks{Corresponding author.}\\
$^1$School of Electronic and Computer Engineering, Peking University, China\\
$^2$NELVT, School of EE\&CS, Peking University, Beijing, China \\
$^3$Nanyang Technological University, Singapore, 
$^4$University of Chinese Academy of Sciences, China\\
$^5$Xiamen University, China, $^6$Peng Cheng Laboratory, China\\
{\tt\small \{ypzhai, shanxb, yhtian\}@pku.edu.cn, shijian.lu@ntu.edu.sg, qxye@ucas.ac.cn,}\\
{\tt\small chenj@pcl.ac.cn, rrji@xmu.edu.cn }
}

\maketitle

\begin{abstract}

Domain adaptive person re-identification (re-ID) is a challenging task, especially when person identities in target domains are unknown. 
Existing methods attempt to address this challenge by transferring image styles or aligning feature distributions across domains, whereas the rich unlabeled samples in target domains are not sufficiently exploited.
This paper presents a novel augmented discriminative clustering (AD-Cluster) technique that estimates and augments person clusters in target domains and enforces the discrimination ability of re-ID models with the augmented clusters.
AD-Cluster is trained by iterative density-based clustering, adaptive sample augmentation, and discriminative feature learning. 
It learns an image generator and a feature encoder which aim to maximize the intra-cluster diversity in the sample space and minimize the intra-cluster distance in the feature space in an adversarial min-max manner.
Finally, AD-Cluster increases the diversity of sample clusters and improves the discrimination capability of re-ID models greatly.
Extensive experiments over Market-1501 and DukeMTMC-reID show that AD-Cluster outperforms the state-of-the-art with large margins.

\end{abstract}

\section{Introduction}

Person re-identification (re-ID) aims to match persons in an image gallery collected from non-overlapping camera networks. 
Despite of the impressive progress of supervised methods 
in person re-ID \cite{chen2019abd} \cite{yang2019attention}, models trained in one domain often fail to generalize well to others due to the change of camera configurations, lighting conditions, person views, etc. Domain adaptive re-ID methods that can work across domains remain a very open research challenge.
%


\begin{figure}[t]
\begin{center}
  \includegraphics[width=1.0\linewidth]{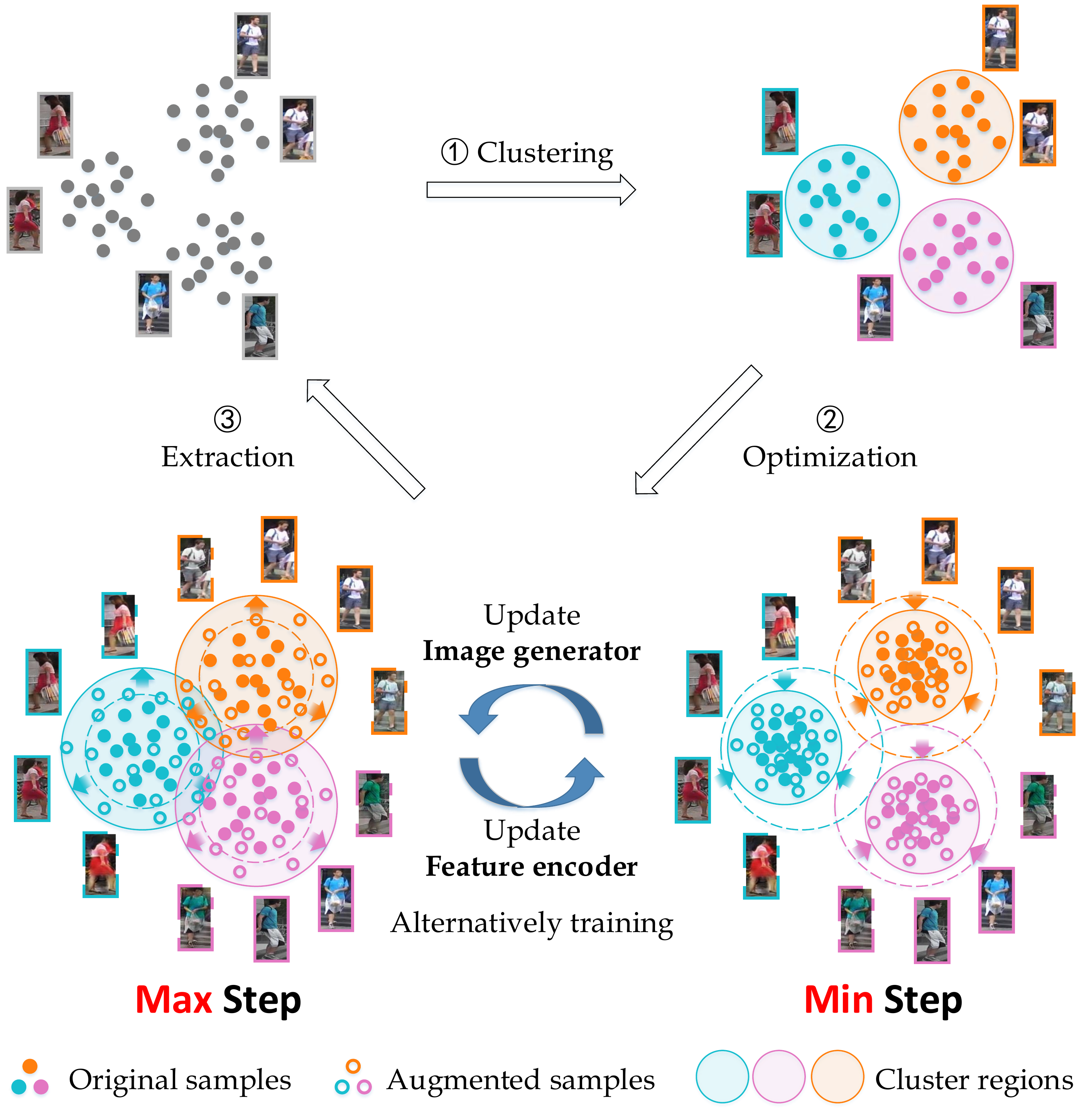}
\end{center}
\caption{AD-Cluster alternatively trains an image generator and a feature encoder, which respectively {\color{red}Max}imizes intra-cluster distance (\textit{i.e.}, increase the diversity of sample space) and {\color{red}Min}imizes intra-cluster distance in feature space (\textit{i.e.}, decrease the distance in new feature space). It enforces the discrimination ability of re-ID models in an adversarial min-max manner. (Best viewed in color)}
\label{fig:motivation}
\end{figure}

To implement domain adaptive re-ID, unsupervised domain adaptation (UDA) methods have been widely explored 
~\cite{Wang_2018_CVPR}, ~\cite{DBLP:conf/bmvc/LinLLK18}, ~\cite{Liu_2019_CVPR}, ~\cite{Deng_2018_CVPR}, ~\cite{Wei_2018_CVPR}, ~\cite{DBLP:conf/eccv/ZhongZLY18}, ~\cite{DBLP:conf/ksem/LvW18}, ~\cite{DBLP:journals/corr/FanZY17}, ~\cite{DBLP:conf/iccv/YeMZLY17}, ~\cite{DBLP:conf/iccv/LiuWL17}.
One major line of UDA methods attempts to align the feature distributions of source and target domains ~\cite{Wang_2018_CVPR}, ~\cite{DBLP:conf/bmvc/LinLLK18}. Another line of methods utilizes adversarial generative models as a style transformer to convert pedestrian images (with identity annotations) of a source domain into a target domain ~\cite{Liu_2019_CVPR}, ~\cite{Deng_2018_CVPR}, ~\cite{Wei_2018_CVPR}, ~\cite{DBLP:conf/ksem/LvW18}. The style-transferred images are then used to train a re-ID model in the target domain. Many UDA methods preserve discriminative information across domains or camera styles, but they largely ignore the unlabeled samples and so the substantial sample distributions in target domains. Recent approaches ~\cite{DBLP:journals/corr/FanZY17}, ~\cite{DBLP:conf/icmcs/WuLLWYL19} alleviate this problem by predicting pseudo-labels in target domains. They leverage the cluster (pseudo) labels for model fine-tuning directly but are often susceptible to noises and hard samples. This prevents them from maximizing model discrimination capacity in target domains.

%

In this paper, we propose an innovative augmented discriminative clustering (AD-Cluster) technique for domain adaptive person re-ID. AD-Cluster aims to maximize model discrimination capacity in the target domain by alternating discriminative clustering and sample generation as illustrated in Fig.\ \ref{fig:motivation}.
%
Specifically, density-based clustering first predicts sample clusters in the target domain where sample features are extracted by a re-ID model that is pre-trained in the source domain. 
%
AD-Cluster then learns through two iterative processes. First, an image generator keeps translating the clustered images to other cameras to augment the training samples while retaining the original pseudo identity labels (i.e. cluster labels).
%
Second, a feature encoder keeps learning to maximize the inter-cluster distance while minimizing the intra-cluster distance in feature space.
The image generator and the feature encoder thus compete in an adversarial min-max manner which iteratively estimate cluster labels and optimize re-ID models. 
Finally, AD-Cluster aggregates the discrimination ability of re-ID models through such adversarial learning and optimization.

The main contributions of this paper can be summarized in three aspects. First, it proposes a novel discriminative clustering method that addresses domain adaptive person re-ID by density-based clustering, adaptive sample augmentation, and discriminative feature learning. Second, it designs an adversarial min-max optimization strategy that increases the intra-cluster diversity and enforces discrimination ability of re-ID models in target domains simultaneously. Third, it achieves significant performance gain over the state-of-the-art on two widely used re-ID datasets: Market-1501 and DukeMTMC-reID.

\begin{figure*}[t]
\begin{center}
  \includegraphics[width=1.0\linewidth]{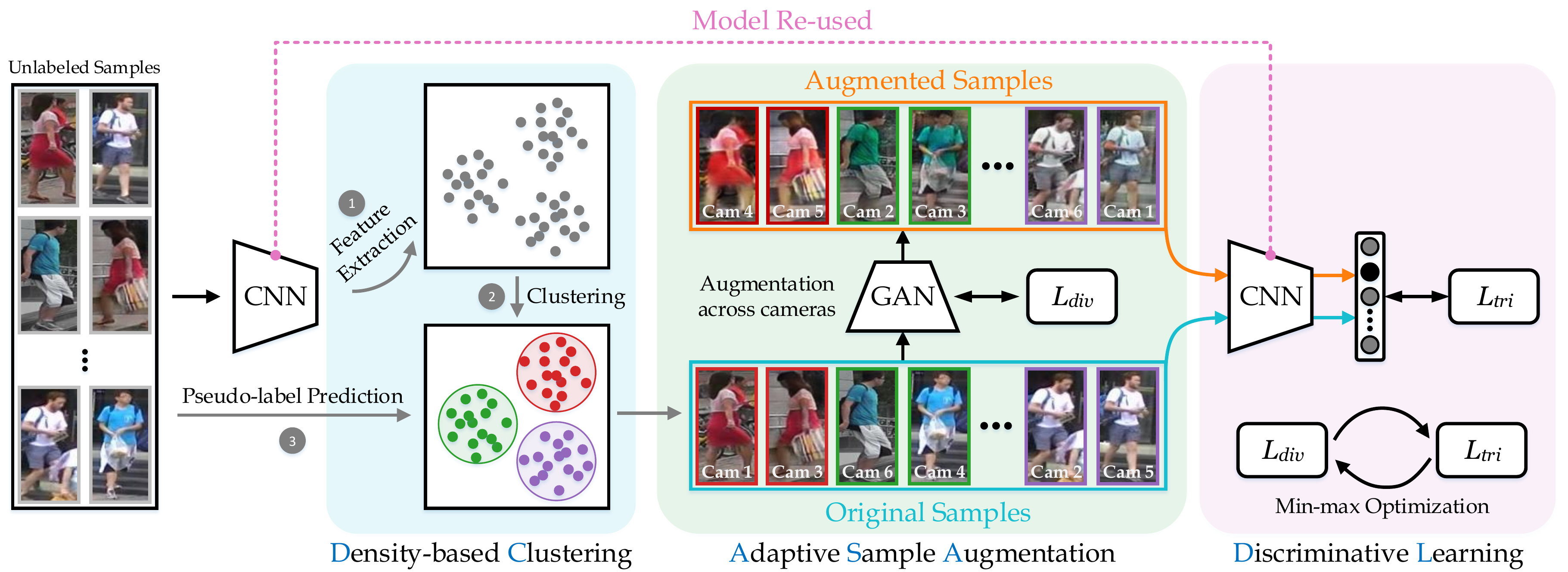}
\end{center}
\caption{The flowchart of the proposed AD-Cluster: The AD-Cluster consists of three components including density-based clustering, adaptive sample augmentation, and discriminative feature learning. Density-based clustering estimates sample pseudo-labels in the target domain. Adaptive sample augmentation maximizes the sample diversity cross cameras while retaining the original pseudo-labels. Discriminative learning drives the feature extractor to minimize the intra-cluster distance. $L_{div}$ denotes the diversity loss and $L_{tri}$ indicates the triplet loss. (Best viewed in color)} 
\label{fig:flowchart}
\end{figure*}
%

\section{Related Works}

While person re-ID has been extensively investigated from various perspectives, we mainly review the domain adaptive person re-ID approaches, which are largely driven by unsupervised domain adaptation (UDA) methods.


\subsection{Unsupervised Domain Adaptation (UDA)}
\textbf{Domain alignment.} 
UDA defines a learning problem where source domains are fully labeled while sample labels in target domains are totally unknown. To learn discriminative modes in target domains, early methods focus on learning feature/sample mapping between source and target domains ~\cite{Saenko2010AdaptingVC},
~\cite{DBLP:conf/aaai/SunFS16}. As an representative method, correlation alignment (CORAL) ~\cite{DBLP:conf/aaai/SunFS16} pursued minimizing domain shift by aligning the mean and co-variance of source and target distributions.
Recent methods ~\cite{DBLP:conf/icml/HoffmanTPZISED18}, ~\cite{DBLP:conf/cvpr/BousmalisSDEK17}, ~\cite{DBLP:conf/nips/LiuT16} attempted reducing the domain shift by using generative adversarial networks (GANs) to learn a pixel-level transformation. The most representative CYCADA ~\cite{DBLP:conf/icml/HoffmanTPZISED18} transferred samples across domains at both pixel- and feature-level.

\textbf{Domain-invariant features.}
The second line of UDA methods focuses on finding domain-invariant feature spaces ~\cite{DBLP:conf/iccv/MotiianPAD17}, ~\cite{DBLP:conf/cvpr/LongD0SGY13}, ~\cite{DBLP:conf/icml/GaninL15}, ~\cite{DBLP:conf/icml/LongC0J15}, ~\cite{DBLP:journals/corr/TzengHZSD14}, ~\cite{DBLP:journals/jmlr/GaninUAGLLML16}, ~\cite{DBLP:journals/corr/AjakanGLLM14}. To fulfill this purpose, Long \emph{et al.} ~\cite{DBLP:conf/icml/LongC0J15}, ~\cite{DBLP:journals/jmlr/GrettonBRSS12} proposed the Maximum Mean Discrepancy (MMD), which maps features of both domains into the same Hilbert space. Ganin \emph{et al.} ~\cite{DBLP:journals/jmlr/GaninUAGLLML16} and Ajakan \emph{et al.} ~\cite{DBLP:journals/corr/AjakanGLLM14} designed domain confusion loss to learn domain-invariant features. Saito \emph{et al.}~\cite{Saito_2018_CVPR} proposed aligning distributions of source and target domains by maximizing the discrepancy of classifiers' outputs.

\textbf{Pseudo-label prediction.}
Another line of UDA methods involves learning representations in target domains by using the predicted pseudo-label. In general, this approach uses an alternative estimation strategy: predicting pseudo-labels of samples by simultaneous modelling and optimizing the model using predicted pseudo-labels ~\cite{DBLP:conf/nips/ChenWB11}, ~\cite{DBLP:conf/nips/RohrbachES13}, ~\cite{DBLP:conf/nips/SenerSSS16}, ~\cite{DBLP:conf/cvpr/ZhangO0018}. In the deep learning era, clustering loss has been designed for CNNs and jointly learning of features, image clusters, and re-ID models in an alternative manner~\cite{DBLP:series/lncs/CoatesN12}, ~\cite{DBLP:conf/cvpr/YangPB16}, ~\cite{DBLP:conf/icml/XieGF16}, ~\cite{DBLP:conf/nips/DosovitskiySRB14}, ~\cite{DBLP:conf/nips/LiaoSZU16}, ~\cite{Caron_2018_ECCV}, ~\cite{Ghasedi_2019_CVPR}.

\subsection{UDA for Person re-ID}
To implement domain adaptive person re-ID, researchers largely referred to the above reviewed UDA methods by incorporating the characteristics of person images. 

\textbf{Domain alignment.}
In~\cite{DBLP:conf/bmvc/LinLLK18}, Lin \emph{et al.} proposed minimizing the distribution variation of the source's and the target's mid-level features based on Maximum Mean Discrepancy (MMD) distance. Wang \emph{et al.}~\cite{Wang_2018_CVPR} utilized additional attribute annotations to align feature distributions of source and target domains in a common space. Other works enforced camera in-variance by learning consistent pairwise similarity distributions~\cite{Wu_2019_ICCV} or reducing the discrepancy between both domains and cameras~\cite{Qi_2019_ICCV}.

GAN-based methods have been extensively explored for domain adaptive person re-ID ~\cite{DBLP:conf/ksem/LvW18}, ~\cite{DBLP:conf/eccv/ZhongZLY18}, ~\cite{Wei_2018_CVPR}, ~\cite{Deng_2018_CVPR}, ~\cite{Liu_2019_CVPR}. 
HHL ~\cite{DBLP:conf/eccv/ZhongZLY18} simultaneously enforced cameras invariance and domain connectedness to improve the generalization
ability of models on the target set. PTGAN~\cite{Wei_2018_CVPR}, SPGAN~\cite{Deng_2018_CVPR}, ATNet~\cite{Liu_2019_CVPR}, CR-GAN~\cite{Chen_2019_ICCV} and PDA-Net~\cite{Li_2019_ICCV} transferred images with identity labels from source into target domains to learn discriminative models. 

By aligning feature and/or appearance, the above methods can preserve well the discriminative information from source domains; however, they largely ignore leveraging the unlabeled samples in target domains, which hinder them from maximizing the model discrimination capacity.

\textbf{Pseudo-label prediction.}
Recently, the problem about how to leverage the large number of unlabeled samples in target domains has attracted increasing attention ~\cite{DBLP:journals/corr/FanZY17}, ~\cite{DBLP:conf/iccv/YeMZLY17}, ~\cite{DBLP:conf/iccv/LiuWL17}, ~\cite{DBLP:conf/icmcs/WuLLWYL19}, ~\cite{Wu_2018_CVPR}, ~\cite{Zhong_2019_CVPR}. Clustering~\cite{DBLP:journals/corr/FanZY17}, ~\cite{DBLP:conf/eccv/ZhengBSWSWT16}, ~\cite{Zhang_2019_ICCV}, ~\cite{Fu_2019_ICCV} and graph matching~\cite{DBLP:conf/iccv/YeMZLY17} methods have been explored to predict pseudo-labels in target domains for discriminative model learning. Reciprocal search~\cite{DBLP:conf/iccv/LiuWL17} and exemplar-invariance approaches~\cite{Wu_2018_CVPR} were proposed to refine pseudo labels, taking camera-invariance into account concurrently. 

Existing approaches have explored cluster distributions in the target domain. On the other hand, they still face the challenge on how to precisely predict the label of hard samples. The hard/difficult samples are crucial to a discriminative re-ID model but they often confuse clustering algorithms. We address this issues by iteratively generating and including diverse and representative samples in the target domain, which enforces the discrimination capability of re-ID models effectively.


\section{The Proposed Approach}


Under the context of unsupervised domain adaptation (UDA) for person re-ID, we have a fully labeled source domain $\{X_s, Y_s\}$ that contains $N_s$ person images of $M$ identities in total in the source domain. $X_s$ and $Y_s$ denote the sample images and identities in the source domain, respectively, where each image $x_{s, i}$ is associated with an identity $y_{s, i}$. In addition, we have an unlabeled target domain $\{X_t\}$ that contains $N_t$ person images. The identities of images in the target domain are unavailable. The goal of AD-Cluster is to learn a re-ID model that generalizes well in the target domain by leveraging labeled samples in the source domain and unlabeled samples in the target domain.

\subsection{Overview}

AD-Cluster consists of two networks including a CNN as the feature encoder $f$ and a Generative Adversarial Network (GAN) as the image generator $g$ as shown in Fig.\ \ref{fig:flowchart}.
The encoder $f$ is first trained using labeled samples in the source domain with cross-entropy loss and triplet loss ~\cite{DBLP:journals/corr/HermansBL17}.
In the target domain, unlabelled sample are represented by features that are extracted by $f$, where density-based clustering groups them to clusters and uses the cluster IDs as the pseudo-labels of the clustered samples. 
With each camera being a new domain with different styles, $g$ translates each sample of the target domain to other cameras and this generates identity-preserving samples with increased diversity.
After that, all samples in the target domain together with those generated are fed to re-train the feature encoder $f$.
The generator $g$ and encoder $f$ thus learn in an adversarial min-max manner iteratively, where $g$ keeps generating identity-preservative samples to maximize the intra-cluster variations in the sample space whereas $f$ learns discriminative representation to minimize the intra-cluster variations in the feature space as illustrated in Fig.\ \ref{fig:motivation}. 

\subsection{UDA Procedure}
\textbf{Supervised learning in source domain:}
In the source domain, the CNN-based person re-ID model is trained by optimizing classification and ranking loss ~\cite{DBLP:journals/corr/HermansBL17}:
\begin{equation}
	\begin{aligned}
	\mathcal{L}_{src} = \mathcal{L}_{cls} + \mathcal{L}_{tri}.
	\end{aligned}
	\label{eq:supervised}
\end{equation}
For a batch of samples, the classification loss is defined by
\begin{equation}
	\begin{aligned}
	\mathcal{L}_{cls} = -\frac{1}{n_s}\sum_{i=1}^{n_s} \log p(y_{s,i}|x_{s,i}),
	\end{aligned}
\end{equation}
where $n_s$, $i$ and $s$ denote the number of images in a batch, image index and source domain, respectively. $p(y_{s,i}|x_{s,i})$ is the predicted probability of image $x_{s,i}$ belonging to $y_{s,i}$.

The ranking triplet loss is defined as
\begin{equation}
	\begin{aligned}
	\mathcal{L}_{tri} = \sum_{i=1}^{n_s} [m &+  \|f(x_{s,i})-f(x_{s,i^{+}})\|_2 \\
	&-  \|f(x_{s,i})-f(x_{s,i^{-}})\|_2],
	\end{aligned}
\label{eq:tripletloss}
\end{equation}
where $x_{s,i^{+}}$ denotes the samples belonging to the same person with $x_{s,i}$. $x_{s,i^{-}}$ denotes the samples belonging to different persons with $x_{s,i}$. $m$ is a margin parameter ~\cite{DBLP:journals/corr/HermansBL17}.


\textbf{Density-based clustering in target domain:}
In each learning iteration, density-based clustering ~\cite{DBLP:conf/kdd/EsterKSX96} is employed in the target domain for pseudo-label prediction. 
The clustering procedure includes three steps: (1) Extracting convolutional features for all person images. (2) Computing a distance matrix with k-reciprocal encoding~\cite{Zhong_2017_CVPR} for all training samples and then performing density-based clustering to assign samples into different groups. (3) Assigning pseudo-labels $Y_t' $ to the training samples $X_t$ according to the groups they belong to.

\textbf{Adaptive sample augmentation across cameras:} \label{section:sa}
Due to the domain gap, the pseudo-labels predicted by density-based clustering suffer from noises. In addition, the limited number of training samples in the target domain often leads to the low diversity of samples in each cluster. These two factors make it difficult to learn discriminative representation in the target domain.

To address these issues, we propose to augment samples in the target domain with a GAN to aggregate sample diversity. The used GAN should possess the following two properties: (1) Generating new person images from existing ones while preserving the original identities; (2) Providing additional invariance such as camera configurations, lighting conditions, and person views. 

To fulfill these purposes, we employ StarGAN ~\cite{Choi_2018_CVPR} to augment person images which can preserve the person identities while generating new images in multiple camera styles. The image generation procedure roots in the results of density-based clustering. Suppose there are $K$ cameras in total in the target domain. 
A StarGAN model is first trained which enables image-image translation between each camera pair. Using the learned StarGAN model, for an image $x_{t,i}$
with pseudo-label $y_{t,i}$, we generate $K$ augmented images $\{ x_{t,i}^{(1)}, y_{t,i}\}, \{x_{t,i}^{(2)}, y_{t,i}\}, ..., \{x_{t,i}^{(K)}, y_{t,i}\}$, which have the pseudo-label $y_{t,i}$ with $x_{t,i}$ and similar styles as the images in camera $1,2, ..., K$, respectively. 
In this way, the sample number in each cluster increases by a factor of $K-1$. 
The augmented images together with original images in target domain are used for discriminative feature learning, according to Eq.\ \ref{eq:tripletloss}.
%
%

\begin{figure}[t]
\begin{center}
  \includegraphics[width=1.0\linewidth]{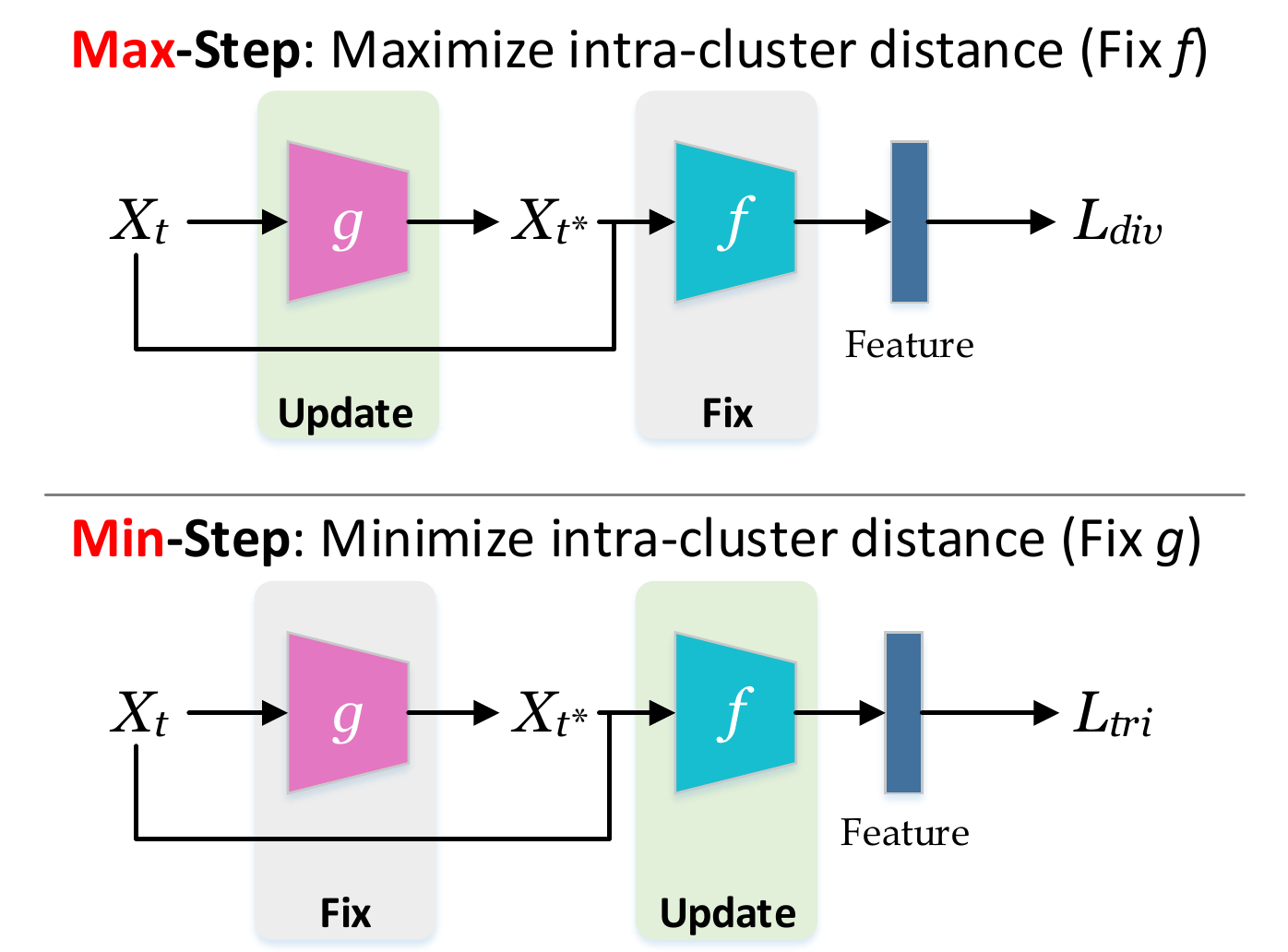}
\end{center}
   \caption{The proposed adversarial min-max learning: With a fixed feature encoder $f$, the generator $g$ learns to generate samples that maximizes intra-cluster distance. With a fixed generator $g$, the feature encoder $f$ learns to minimize the intra-cluster distance and maximize the inter-cluster distance under the guide of triplet loss.} 
\label{fig:opt}
\end{figure}

\subsection{Min-Max Optimization}

Although the adaptive sample augmentation enforces the discrimination ability of re-ID models,
the sample generation procedure is completely independent from the clustering and feature learning which could lead to insufficient sample diversity across cameras.

To fuse the adaptive data augmentation with discriminative feature learning, we propose an adversarial min-max optimization strategy as illustrated in Fig.~\ref{fig:opt}. Specifically, we alternatively train an image generator and a feature encoder that maximize sample diversity and minimize intra-cluster distance for each mini-batch, respectively.

\textbf{Max-Step:} Star-GAN ~\cite{Choi_2018_CVPR} is employed as an image generator ($g$) for a given feature encoder ($f$). 
In the procedure, the summation of Euclidean distances between samples and their cluster centers is defined as cluster diversity $\mathcal{D}_{div}$. For each sample, the diversity is defined as
\begin{equation}
	\begin{aligned}
\mathcal{D}_{div}(x_{t,i}) &=  \left\| {f(g(x_{t,i}))-\frac{1}{\sum_{j=1}^{n_t} a(i,j)} \sum_{j=1}^{n_t} a(i,j)f(x_{t,i})} \right\|_2, \\
	\end{aligned}
	\label{eq:diversity}
\end{equation}
where $a(i,j)$ indicates whether sample $x_{t,i}$ and $x_{t,j}$ belong to the same person or not. $a(i,j)=1$ when $y_{t,i}=y_{t,j}$, otherwise $a(i,j)=0$.

For a batch of sample, a diversity loss is defined as
\begin{equation}
	\begin{aligned}
	\mathcal{L}_{div} &= \frac{1}{n_t} \sum_{i=1}^{n_t} e^{-\lambda \mathcal{D}_{div}(x_{t,i})} ,\\
	\end{aligned}
\label{eq:diversitiy_loss}
\end{equation}
where $\lambda$ is hyper-parameter. We use a negative exponent function to prevent $\mathcal{D}_{div}$ from growing too large so as to preserve the identity of the augmented person images. According to Eq.\ \ref{eq:diversity} and Eq.\ \ref{eq:diversitiy_loss}, maximizing the sample diversity $\mathcal{D}_{div}$ in a cluster is equal to minimizing the loss, as
\begin{equation}
\mathop{\arg\max}_{g} \mathcal{D}_{div} \Leftrightarrow \mathop{\arg\min}_{g} \mathcal{L}_{div}.
\end{equation}

$\mathcal{L}_{div}$ is combined with loss of StarGAN to optimize the generator $g$ while augmenting samples. 

\begin{figure}[t]
\begin{center}
  \includegraphics[width=1.0\linewidth]{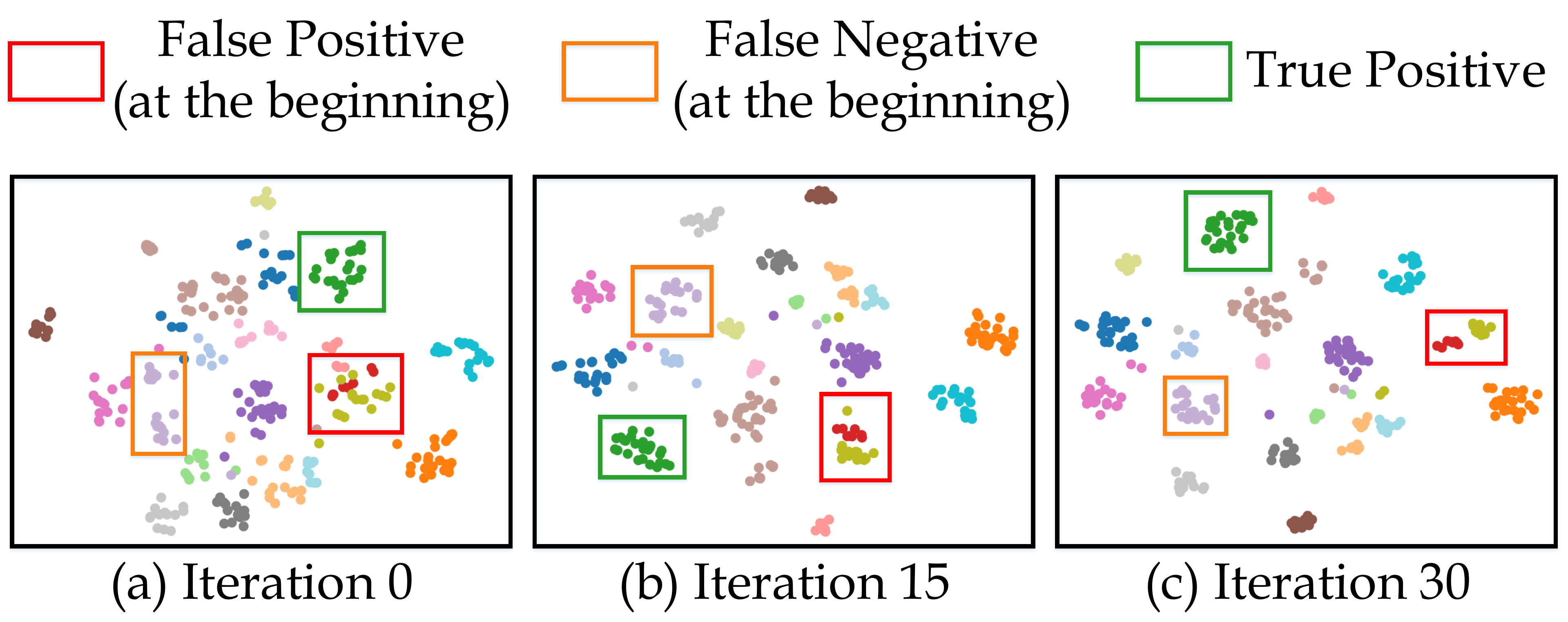}

\end{center}
  \vspace{-0.2cm}
  \caption{The sparsely and incorrectly distributed person image features of different identities are grouped to more compact and correct clusters through the iterative clustering process. (Best viewed in color with zoom in.)} 
\label{fig:group}
\vspace{-0.4cm}
\end{figure}

\textbf{Min-Step:} Given a fixed generator $g$, the feature encoder $f$ learns to minimize the intra-cluster distance while maximizing inter-cluster distance in feature space under the constraint of triplet loss, which is defined as
\begin{equation}
	\begin{aligned}
	\mathcal{L}_{tri} = \sum_{i=1}^{n_t} [m &+  \|f(x_{t,i})-f(x_{t,i^{+}})\|_2 \\
	&-  \|f(x_{t,i})-f(x_{t,i^{-}})\|_2],
	\end{aligned}
\label{eq:min_tripletloss}
\end{equation}
where $x_{t,i^{+}}$ denotes the samples belonging to the same cluster with $x_{t,i}$. $x_{t,i^{-}}$ denotes the samples belonging to different clusters with $x_{t,i}$. $m$ is a margin parameter. Specifically, we choose all the positive samples and the hardest negative sample to construct the triplets for each anchor sample, with a mini-batch of both original and generated sample images.
The objective function is defined by
\begin{equation}
	\begin{aligned}
	\mathop{\arg\min}_{f} \; \mathcal{D}_{div} \Leftrightarrow \mathop{\arg\min}_{f} \mathcal{L}_{tri}.
	\end{aligned}
\end{equation}
When $g$ keeps producing more diverse samples with features far away from the cluster centers, $f$ will be equipped with stronger discrimination ability in the target domain, as illustrated in Fig. \ref{fig:group}. 
\textbf{Algorithm \ref{alg:algorithm}} shows the detailed training procedure of the proposed AD-Cluster.

\begin{algorithm}[t]
\caption{Training procedure of AD-Cluster}
\label{alg:algorithm}
\textbf{Input}:  Source domain dataset $\mathbf{S}$, target domain dataset $\mathbf{T}$\\
\textbf{Output}: Feature encoder $f$
\begin{algorithmic}[1] 
\STATE Pre-train feature encoder $f$ on $\mathbf{S}$ by optimizing Eq. \ref{eq:supervised}. 
\FOR{each clustering iteration}
\STATE Extract features $\mathbf{F}=f(\mathbf{T})$.
\STATE Cluster training samples in target domain using $\mathbf{F}$.
\FOR{each mini-batch $\mathcal{B}\subset \mathbf{T}$}
\STATE Max-step: train image generator $g$ by $\mathcal{B}$.
\STATE Min-step: train feature encoder $f$ by $\{\mathcal{B},g(\mathcal{B})\}$.
\ENDFOR
\ENDFOR
\STATE \textbf{return} Feature encoder $f$
\end{algorithmic}
\end{algorithm}

\begin{table*}[t] 
\begin{center}
\begin{tabular}{l|p{1.1cm}<{\centering}p{1.1cm}<{\centering}p{1.1cm}<{\centering}p{1.1cm}<{\centering}|p{1.1cm}<{\centering}p{1.1cm}<{\centering}p{1.1cm}<{\centering}p{1.1cm}<{\centering}}
    \hline
    \multirow{2}{*}{Methods} & \multicolumn{4}{c|}{DukeMTMC-reID $\to$ Market-1501} & \multicolumn{4}{c}{Market-1501 $\to$ DukeMTMC-reID} \\
    \cline{2-9}
    & R-1 & R-5 & R-10 & mAP & R-1 & R-5 & R-10 & mAP \\
    \hline\hline
    LOMO~\cite{Liao_2015_CVPR}             & 27.2 & 41.6 & 49.1 & 8.0  & 12.3 & 21.3 & 26.6 & 4.8  \\
    Bow~\cite{Zheng_2015_ICCV}               & 35.8 & 52.4 & 60.3 & 14.8 & 17.1 & 28.8 & 34.9 & 8.3  \\
    \hline
    UMDL~\cite{Peng_2016_CVPR}              & 34.5 & 52.6 & 59.6 & 12.4 & 18.5 & 31.4 & 37.6 & 7.3  \\
    PTGAN~\cite{Wei_2018_CVPR}             & 38.6 & -    & 66.1 & -    & 27.4 & -    & 50.7 & -    \\
    PUL~\cite{DBLP:journals/tomccap/FanZYY18}               & 45.5 & 60.7 & 66.7 & 20.5 & 30.0 & 43.4 & 48.5 & 16.4 \\
    SPGAN~\cite{Deng_2018_CVPR}             & 51.5 & 70.1 & 76.8 & 22.8 & 41.1 & 56.6 & 63.0 & 22.3 \\
    CAMEL~\cite{Yu_2017_ICCV}             & 54.5 & -    & -    & 26.3 & -    & -    & -    & -    \\
    ATNet~\cite{Liu_2019_CVPR}             & 55.7 & 73.2 & 79.4 & 25.6 & 45.1 & 59.5 & 64.2 & 24.9 \\
    MMFA~\cite{DBLP:conf/bmvc/LinLLK18}              & 56.7 & 75.0 & 81.8 & 27.4 & 45.3 & 59.8 & 66.3 & 24.7 \\
    SPGAN+LMP~\cite{Deng_2018_CVPR}         & 57.7 & 75.8 & 82.4 & 26.7 & 46.4 & 62.3 & 68.0 & 26.2 \\
    TJ-AIDL~\cite{Wang_2018_CVPR}           & 58.2 & 74.8 & 81.1 & 26.5 & 44.3 & 59.6 & 65.0 & 23.0 \\
    CamStyle~\cite{DBLP:journals/tip/ZhongZZLY19}          & 58.8 & 78.2 & 84.3 & 27.4 & 48.4 & 62.5 & 68.9 & 25.1 \\
    HHL~\cite{DBLP:conf/eccv/ZhongZLY18}               & 62.2 & 78.8 & 84.0 & 31.4 & 46.9 & 61.0 & 66.7 & 27.2 \\
    ECN~\cite{Zhong_2019_CVPR}               & \underline{75.1} & \underline{87.6} & 
                        \underline{91.6} & \underline{43.0} & 
                        \underline{63.3} & \underline{75.8} & 
                        \underline{80.4} & \underline{40.4} \\
    UDAP~\cite{DBLP:journals/corr/abs-1807-11334}              & \emph{75.8} & \emph{89.5} & 
                        \emph{93.2} & \emph{53.7} & 
                        \emph{68.4} & \emph{80.1} & 
                        \emph{83.5} & \emph{49.0} \\
    \hline
    AD-Cluster (Ours) & \textbf{86.7} & \textbf{94.4} & 
                        \textbf{96.5} & \textbf{68.3} & 
                        \textbf{72.6} & \textbf{82.5} & 
                        \textbf{85.5} & \textbf{54.1} \\
\hline
\end{tabular}
\end{center}
\caption{Comparison of the proposed AD-Cluster with state-of-the-art methods: For the transfers DukeMTMC-reID $\rightarrow$ Market-1501 and Market-1501 $\rightarrow$ DukeMTMC-reID, the proposed AD-Cluster significantly outperforms all state-of-the-art methods over all evaluation metrics. The top-three results are highlighted with bold, italic, and underline fonts, respectively.}
\label{table:staart}
\end{table*}

\section{Experiments}

We detail the implementation and evaluation of AD-Cluster. During the evaluation, ablation studies, parameter analysis, and comparisons with other methods are provided.

\subsection{Datasets and Evaluation Metrics}
The experiments were conducted over two public datasets Market1501~\cite{Zheng_2015_ICCV} and DukeMTMC-ReID~\cite{DBLP:conf/eccv/RistaniSZCT16}~\cite{Zheng_2017_ICCV}
by using the evaluation metrics Cumulative Matching Characteristic (CMC) curve and mean average precision (mAP).

\textbf{Market1501}~\cite{Zheng_2015_ICCV}: This dataset contains 32,668 images of 1,501 identities from 6 disjoint surveillance cameras. Of the 32,668 person images, 12,936 images from 751 identities form a training set, 19,732 images from 750 identities (plus a number of distractors) form a gallery set, and 3,368 images from 750 identities form a query set.


\textbf{DukeMTMC-ReID}~\cite{DBLP:conf/eccv/RistaniSZCT16}~\cite{Zheng_2017_ICCV}: This dataset is a subset of the DukeMTMC. It consists of 16,522 training images, 2,228 query images, and 17,661 gallery images of 1,812 identities captured using 8 cameras. Of the 1812 identities, 1,404 appear in at least two cameras and the rest 408 (considered as distractors) appear in only one camera.



\subsection{Implementation Details} We adopt the ResNet-50 ~\cite{He_2016_CVPR} as the backbone network and initialize it by using parameters pre-trained on the ImageNet ~\cite{DBLP:conf/cvpr/DengDSLL009}. During training, the input image is uniformly resized to $256\times128$ and traditional image augmentation is performed via random flipping and random erasing. 
For each identity from the training set, a mini-batch of size 256 is sampled with P = 32 randomly selected identities and K = 8 (original to augmented samples ratio = 3:1) randomly sampled images for computing the hard batch triplet loss. 

In addition, we set the margin parameter at 0.5 and use the SGD optimizer to train the model. The learning rate is set at $6\times10^{-5}$ and momentum at $0.9$. The whole training process consists of 30 iterative min-max clustering process, each of which consists of 70 training epochs.

Our network was implemented on a PyTorch platform and trained using 4 NVIDIA Tesla K80 GPUs (each with 12GB VRAM). 

\subsection{Comparisons with State-of-the-Arts} 
We compare AD-Cluster with state-of-the-art unsupervised person ReID methods including: 1) LOMO~\cite{Liao_2015_CVPR} and BOW~\cite{Zheng_2015_ICCV} that used hand-crafted features; 2) UMDL~\cite{Peng_2016_CVPR}, PUL~\cite{DBLP:journals/tomccap/FanZYY18} and CAMEL~\cite{Yu_2017_ICCV} that employed unsupervised learning; 
and 3) nine UDA-based methods including PTGAN~\cite{Wei_2018_CVPR}, SPGAN~\cite{Deng_2018_CVPR}, ATNet~\cite{Liu_2019_CVPR}, CamStyle~\cite{DBLP:journals/tip/ZhongZZLY19}, HHL~\cite{DBLP:conf/eccv/ZhongZLY18}, and ECN~\cite{Zhong_2019_CVPR} that used GANs; MMFA~\cite{DBLP:conf/bmvc/LinLLK18} and TJ-AIDL~\cite{Wang_2018_CVPR} that used images attributes; and UDAP~\cite{DBLP:journals/corr/abs-1807-11334} that employed clustering. 
Table \ref{table:staart} shows the person Re-ID performance while adapting from Market1501 to DukeMTMC-reID and vice versa.


As Table \ref{table:staart} shows, LOMO and BOW using hand-crafted features do not perform well. UMDL~\cite{Peng_2016_CVPR}, PUL~\cite{DBLP:journals/tomccap/FanZYY18} and CAMEL~\cite{Yu_2017_ICCV} derive image features through unsupervised learning, and they perform clearly better than LOMO and BOW under most evaluation metrics. The UDA-based methods further improve the person Re-ID performance in most cases. Specifically, UDAP performs much better than other methods as it employed the distribution of clusters in the target domains.
The performance of the UDA methods using GAN is diverse. In particular, ECN performs better than most methods using GANs because it enforces cameras invariance and domain connectedness.  

\begin{table*}[t]
\begin{center}
\begin{tabular}{l|p{1.1cm}<{\centering}p{1.1cm}<{\centering}p{1.1cm}<{\centering}p{1.1cm}<{\centering}|p{1.1cm}<{\centering}p{1.1cm}<{\centering}p{1.1cm}<{\centering}p{1.1cm}<{\centering}}
    \hline
    \multirow{2}{*}{Methods} & \multicolumn{4}{c|}{DukeMTMC-reID $\to$ Market-1501} & \multicolumn{4}{c}{Market-1501 $\to$ DukeMTMC-reID} \\
    \cline{2-9}
    & R-1 & R-5 & R-10 & mAP & R-1 & R-5 & R-10 & mAP \\
    \hline\hline
    Supervised Model (upper bound)      & 91.9 & 97.4 & 98.4 & 81.4  & 82.8 & 92.2 & 94.9 & 69.8  \\ 
    \hline
    Direct Transfer          & 46.3 & 63.8 & 71.2 & 21.3  & 28.0 & 42.9 & 49.4 & 14.2  \\
    Baseline                       & 73.8 & 85.7 & 89.0 & 51.0  & 68.6 & 79.3 & 82.2 & 49.0  \\
    Baseline+ASA                    & 83.3 & 93.6 & 95.7 & 62.8  & 71.5 & 81.1 & 84.2 & 52.7  \\
    Baseline+ASA+DL                 & 86.7 & 94.4 & 96.5 & 68.3  & 72.6 & 82.5 & 85.5 & 54.1  \\ 
    \hline
\end{tabular}
\end{center}
\caption{Ablation studies of AD-Cluster: \textit{Supervised Models:} Re-ID models trained by using the labelled training images of the target domain; \textit{Direct Transfer:} Re-ID models trained by using the labelled training images of the source domain; \textit{Baseline:} Baseline Re-ID models trained via Density-based Clustering ~\cite{DBLP:conf/kdd/EsterKSX96}; \textit{Baseline+ASA:} Baseline model plus the proposed Adaptive Sample Augmentation; \textit{Baseline+ASA+DL:} Baseline model plus the proposed Sample Augmentation and Discriminative Feature Learning.} 
\label{table:ablation}
\end{table*}

In addition, AD-Cluster performs significantly better than all compared methods. As Table \ref{table:staart} shows, AD-Cluster achieves a rank-1 accuracy of $86.7\%$ and an mAP of $68.3\%$ for the unsupervised adaptation DukeMTMC-reID $\rightarrow$ Market1501, which outperforms the state-of-the-art (by UDAP) by $10.9\%$ 
and $14.6\%$, respectively. 
For Market1501 $\rightarrow$ DukeMTMC-reID, AD-Cluster obtains a rank-1 accuracy of $72.6\%$ and an mAP of $54.1\%$ which outperforms the state-of-the-art (by UDAP) by $4.2\%$ and $5.1\%$, respectively. 

\begin{figure*}[t]
\begin{center}
  \includegraphics[width=1.0\linewidth]{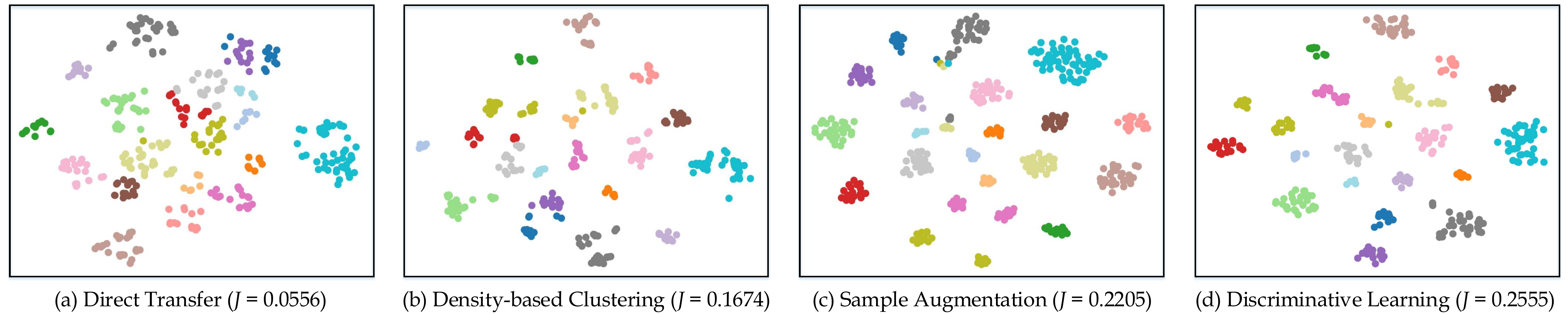}
\end{center}
   \caption{Comparison of sample distributions on Market-1501 dataset with different transfer techniques: $J$ denotes the ratio between inter-class scatter and intra-class scatter and a larger $J$ means better transfer. (Best viewed in color)} 
\label{fig:cmp_dis}
\end{figure*}

Note that AD-Cluster improves differently for the two adaptations in reverse directions between the two datasets. This can also be observed for most existing methods as shown in Table \ref{table:staart}.
We conjecture that this is because the large variance of samples in DukeMTMC-reID caused more clustering noise, which reduces the effectiveness of pseudo-label prediction and hinders the model adaptation. 

\subsection{Ablation Studies}
Extensive ablation studies are performed to evaluate each component of AD-Cluster as shown in Table \ref{table:ablation}.

\textbf{Baseline, the Upper and Lower Bounds: }
We first derive the upper and lower performance bounds for the ablation studies as shown in Table \ref{table:ablation}. Specifically, the upper bounds of Re-ID performance are derived by the \textit{Supervised Models} which are trained by using labelled target-domain training images and evaluated over the target-domain test images. The lower performance bounds are derived by the \textit{Direct Transfer} models which are trained by using the labelled source-domain training images and evaluated over the target-domain test images. We can observe huge performance gaps between the \textit{Direct Transfer} models and the \textit{Supervised Models} due to the domain shift. Take the Market-1501 as an example. The rank-1 accuracy of the supervised model reaches up to $91.9\%$ but it drops significantly to $46.3\%$ for the directly transferred model which is trained by using the DukeMTMC-reID training images. 

In addition, Table \ref{table:ablation} gives the performance of \textit{Baseline} models which are transfer models as trained by iterative density-based clustering as described in ~\cite{DBLP:journals/corr/abs-1807-11334}. As Table \ref{table:ablation} shows, the 
\textit{Baseline} model outperforms the \textit{Direct Transfer} model by a large margin. For example, the rank-1 accuracy improves from $46.3\%$ to $73.8\%$ and from $28.0\%$ to $68.6\%$, respectively, while evaluated over the datasets Market1501 and DukeMTMC-reID. This shows that the density-based clustering in the \textit{Baseline} can group samples of same identities to any irregular distributions by utilizing the density correlation. At the same time, we can observe that there are still large performance gaps between the \textit{Baseline} models and the \textit{Supervised Models}, $e.g.,$ a drop of $30\%$ in mAP while transferring from DukeMTMC-reID to Market1501.


\textbf{Adaptive Sample Augmentation:} We first evaluated the adaptive sample augmentation as described in Section \ref{section:sa}. For this experiment, we designed a network \textit{Baseline+ASA} that just incorporates the adaptive sample augmentation into the \textit{Baseline} that performs transfer via iterative density-based clustering. As shown in Table \ref{table:ablation}, adaptive sample augmentation improves the re-ID performance significantly. For DukeMTMC-reID $\rightarrow$ Market1501, the \textit{Baseline+ASA} achieves a rank-1 accuracy of $83.3\%$ and an mAP of $62.8\%$ which are higher than the \textit{Baseline} by $9.5\%$ and $11.8\%$, respectively. The contribution of the proposed sample augmentation can also be observed in the perspective of sample distributions in the feature space as illustrated in Fig.~\ref{fig:cmp_dis}(c), where the including of the proposed sample augmentation improves the sample distribution greatly as compared with density-based clustering as shown in Fig.~\ref{fig:cmp_dis}(b).

The large performance improvements can be explained by the effectiveness of the augmented samples. Specifically, the iterative injection of ID-preserving cross-camera images helps to reduce the feature distances of person images within the same cluster ($i.e.$, the intra-cluster distances) and increase that of different clusters ($i.e.$, the  inter-cluster distances) simultaneously.

\textbf{Discriminative Learning:} We evaluated the the discriminative learning component as described in Section 3.3. For this experiment, we designed a new network \textit{Baseline+ASA+DL} that further incorporates discriminative learning into the \textit{Baseline+ASA} network as described in the previous subsection. As shown in Table \ref{table:ablation}, the incorporation of discriminative learning consistently improves the person Re-ID performance beyond the \textit{Baseline+ASA}. Take the transfer DukeMTMC-reID $\rightarrow$ Market1501 as an example. The \textit{Baseline+ASA+DL} achieves a rank-1 accuracy of $86.7\%$ and an mAP of $68.3\%$ which outperforms the corresponding \textit{Baseline+ASA} by $3.4\%$ and $5.5\%$, respectively. The superior performance of the proposed discriminative learning can also be observed intuitively in the perspective of sample distributions in feature space as shown in Fig.~\ref{fig:cmp_dis}(d). The effectiveness of the discriminative learning can be largely attributed to the min-max clustering optimization that alternately trains the image generator to generate more diverse samples for maximizing the sample diversity and the feature encoder for minimizing the intra-class distance.

From another perspective, it can be seen that \textit{Baseline+ASA+DL} ($i.e.$, the complete AD-Cluster model) outperforms the \textit{Baseline} by up to $13\%$ in rank-1 accuracy and 17\% in mAP, respectively. 
This demonstrates the effectiveness of the proposed ID-preserving cross-camera sample augmentation and discriminative learning in UDA-based person Re-ID. 
In addition, we can observe that the performance of \textit{Baseline+ASA+DL} becomes even close to the \textit{Supervised Models}. For example, the \textit{Baseline+ASA+DL} achieves a rank-1 accuracy of 86.7\% for the transfer DukeMTMC-reID $\rightarrow$ Market-1501 which is only 5.2\% lower than the corresponding Supervised Model.

\begin{figure}[t]
\begin{center}
  \includegraphics[width=1.0\linewidth]{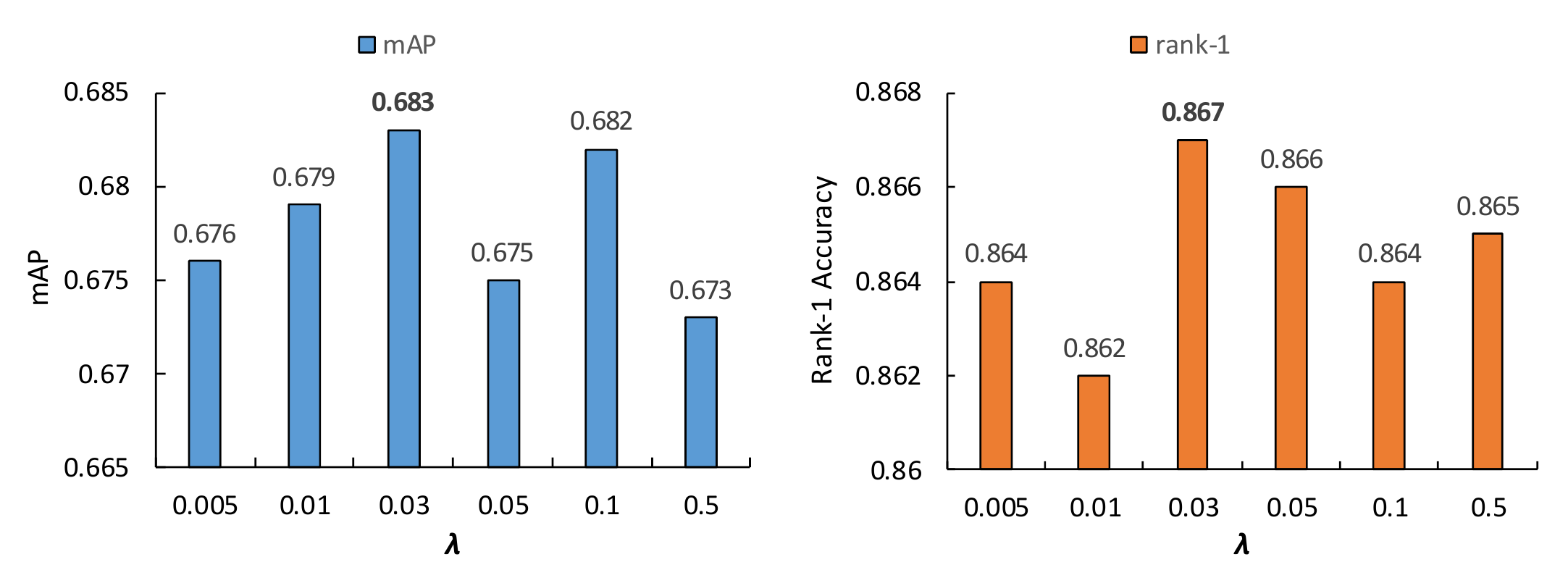}
\end{center}
\caption{The min-max attenuation coefficient $\lambda$ in Eq.\ \ref{eq:diversitiy_loss} affects both mAP and rank-1 accuracy (evaluated on Market-1501).} 
\label{fig:lambda}
\vspace{-0.4cm}
\end{figure}

\textbf{Specificity of AD-Cluster.}
The performance of the AD-Cluster is related to the sample generation method. In this work, we generate cross-camera images by using Star-GAN which theoretically can be replaced by any other ID-preserving generators. The key is how well the re-ID model can learn camera style in-variance via generating new samples. The AD-Cluster could thus be influenced by two factors: the quality of generated samples and the strength of camera style in-variance of the sample distribution in the target domain. These variances explain the different improvements by AD-Cluster over different adaptation tasks.




\subsection{Discussion}
%
The min-max attenuation coefficient $\lambda$ in Eq.\ \ref{eq:diversitiy_loss} will affect the ID-preserving min-max clustering and so the person Re-ID performance. 
We studied this parameter by setting it to different values and checking the person Re-ID performance. 
Fig.  \ref{fig:lambda} shows experimental results on Market-1501. Using a smaller $\lambda$ usually leads to a higher cluster diversity, which further leads to better Re-ID performance. On the other hand, $\lambda$ should not be very small for the target of identity preservation. Experiments show that AD-Cluster performs best when $\lambda=0.03$.
\begin{figure}[t]
\begin{center}
  \includegraphics[width=1.0\linewidth]{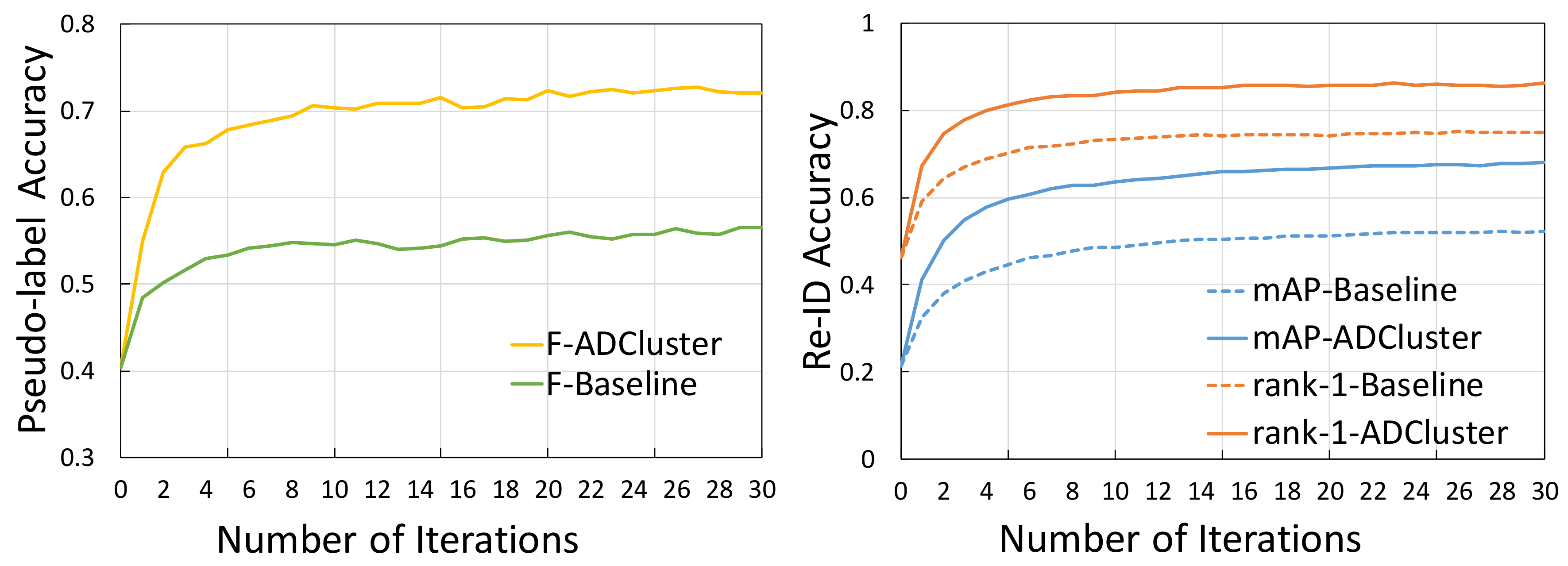}
\end{center}
  \caption{Iterative min-max clustering outperforms density-based clustering consistently for both accuracy of pseudo-label prediction on the left and mAP \& rank-1 accuracy of person Re-ID on the right (for DukeMTMC-reID $\rightarrow$ Market1501).} 
\label{fig:iteration}
\vspace{-0.2cm}
\end{figure}
We also evaluate the accuracy of the pseudo-labels that are predicted during the iterative min-max clustering, as well as how the person Re-ID performance evolves during this process. 
Fig. \ref{fig:iteration} (left) shows that the f-score of the predicted pseudo-labels keeps improving during the iterative clustering process. Additionally, the proposed min-max clustering outperforms the density-based clustering~\cite{DBLP:conf/kdd/EsterKSX96} significantly in both mAP and rank-1 accuracy as shown in the right graph in Fig. \ref{fig:iteration}.

\section{Conclusion}
This paper presents an augmented discriminative clustering (AD-Cluster) method for domain adaptive person re-ID. With density-based clustering, we introduce adaptive sample augmentation to generate more diverse samples and a min-max optimization scheme to learn more discriminative re-ID model. Experiments demonstrates the effectiveness of adaptive sample augmentation and min-max optimization for improving the discrimination ability of deep re-ID model. Our approach not only produces a new state-of-the-art in UDA accuracy on two large-scale benchmarks but also provides a fresh insight for general UDA problems. We expect that the proposed AD-Cluster will inspire new insights and attract more interests for better UDA-based recognition \cite{Fu_2019_ICCV} and detection \cite{zhang2019freeanchor} in the near future.

\section*{Acknowledgement}
This work is partially supported by grants from the National Key R\&D Program of China under grant 2017YFB1002400, the National Natural Science Foundation of China under contract No. 61825101, No. U1611461, No. 61836012 and No. 61972217.

{\small
\bibliographystyle{ieee_fullname}
\bibliography{egbib}
}

\end{document}